\documentclass[conference,compsoc]{IEEEtran}

\ifCLASSOPTIONcompsoc
  \usepackage[nocompress]{cite}
\else
  \usepackage{cite}
\fi

\usepackage{graphicx}

\usepackage{marvosym}

\usepackage{multirow}
\usepackage{makecell}
\usepackage{booktabs}
\usepackage{arydshln}

\usepackage{rotating}

\usepackage{amssymb}
\newcommand\rot[1]{\rlap{\rotatebox{90}{#1}}}
\newcommand\OK{$\checkmark$}

\IEEEoverridecommandlockouts

\begin{document}

\title{Extensive Threat Analysis of Vein Attack Databases and Attack Detection by Fusion of Comparison Scores}

\author{Johannes~Schuiki,
        Michael~Linortner,
        Georg~Wimmer,
        and~Andreas~Uhl

        \thanks{This is a preprint of the following chapter: Schuiki, J., Linortner, M., Wimmer, G., Uhl, A., Extensive Threat Analysis of Vein Attack Databases and Attack Detection by Fusion of Comparison Scores, published in Handbook of Biometric Anti-Spoofing Third Edition: Presentation Attack Detection and Vulnerability Assessment, edited by Marcel, S., Fierrez, J., Evans, N., 2023, Springer, Singapore reproduced with permission of Springer Nature Singapore Pte Ltd. The final authenticated version is available online at: https://doi.org/10.1007/978-981-19-5288-3\_17}
        }

\maketitle

\begin{abstract}
The last decade has brought forward many great contributions regarding presentation attack detection for the domain of finger and hand vein biometrics. Among those contributions, one is able to find a variety of different attack databases that are either private or made publicly available to the research community. However, it is not always shown whether the used attack samples hold the capability to actually deceive a realistic vein recognition system. Inspired by previous works, this study provides a systematic threat evaluation including three publicly available finger vein attack databases and one private dorsal hand vein database. To do so, 14 distinct vein recognition schemes are confronted with attack samples and the percentage of wrongly accepted attack samples is then reported as the Impostor Attack Presentation Match Rate. As a second step, comparison scores from different recognition schemes are combined using score level fusion with the goal of performing presentation attack detection.
\end{abstract}

\IEEEdisplaynontitleabstractindextext
\IEEEpeerreviewmaketitle

\section{Introduction}
\label{sec:introduction}

Since most biometric data is inseparably linked to a human individual, security considerations for biometric systems are of utter importance. However, as some biometric traits are left behind (e.g. latent fingerprint) or are always exposed to public (e.g. gait, face), one potential attack scenario that has to be dealt with is the presentation of a replica of a biometric sample to the biometric reader with the goal of either impersonate someone or not being recognized. This is known as presentation attack and defined by the ISO/IEC \cite{iso_30107} as: \textit{presentation to the biometric data capture subsystem with the goal of interfering with the operation of the biometric system}.

Due to the circumstance that blood vessels are hard to detect with the human visual system or consumer cameras, biometric systems that employ vascular patterns in the hand region were often attributed with being insusceptible against such attacks. However, German hackers \cite{ccc18a} demonstrated the use of a hand vein pattern which was acquired from a distance of a few meters to successfully deceive a commercial hand vein recognition device during a live session at a hacking conference in 2018. Besides that, they suggested that hand dryers in the restroom could be manipulated to secretly capture hand vein images. Also one must never rule out the possibility of an already existing vein database being compromised by an attacker. Hence, plenty of studies were published that address the possibility of vein recognition systems getting deceived by forged samples over the course of the last decade. A comprehensive overview of countermeasures to such attacks (also known as presentation attack detection, \textit{PAD}) can be found in table 14.1 in \cite{Kolberg2020}. 
\\

In order to evaluate the effectiveness of such PAD algorithms, either private databases or publicly available databases are used. For the creation of such attack samples, the following strategies can be found in literature:

One of the pioneer works was published in 2013 by Nguyen et al. \cite{Nguyen2013}, where one of the first successful attempts to fool a finger vein recognition algorithm was reported. They created a small scale attack database from an existing finger vein database consisting of seven individuals. Their recipe for generating the presentation attacks was to print selected finger vein templates from an existing database onto two types of paper and on overhead projector film using a laser printer, thereby creating three different attack types.  

Tome et al. \cite{PTome2014a} created presentation attacks using a subset from a publicly available finger vein recognition database, that was initially released at the same time. Likewise, as Nguyen et al. in 2013 \cite{Nguyen2013}, a laser printer was used for presentation attack creation and additionally the contours of the veins were enhanced using a black whiteboard marker. One year later, in 2015, the first competition on counter measures to finger vein spoofing attacks was held \cite{PTome2015a} by the same authors as in \cite{PTome2014a}. With this publication, their presentation attack database was extended such that every biometric sample from their reference database has a corresponding spoofed counterpart. This was the first complete finger vein presentation attack database publicly available for research purposes. 

Instead of merely printing vein samples on paper, Qiu et al. \cite{Qiu2018a} suggested to print the sample on two overhead projector films and sandwich a white paper in between the aligned overhead films. The white paper is meant to reduce overexposure due to transparency of the overhead films.

Another experiment was shown in \cite{Patil2016}. Here dorsal hand vein images were acquired using a smartphone camera that had, without modification, a moderate sensitivity in the near infrared region. The images were then shown on the smartphone display and presented to the biometric sensor. 

A very different approach was published by Otsuka et al. \cite{Otsuka2016} in 2016. They created a working example of what they call 'wolf attack'. This type of attack has the goal to exploit the working of a recognition toolchain to construct an attack sample that will generate a high similarity score with any biometric template stored. They showed that their master sample worked in most of the finger vein verification attempts, therefore posing a threat to this particular recognition toolchain used. 

In 2020, Debiasi et al. \cite{Debiasi20a} created a variety of presentation attacks that employ wax and silicone casts. However, no successful comparison experiments with bona fide samples were reported. In \cite{Schuiki21a}, the attack generation recipe using casts of beeswax was reworked and finally shown to be functional. 

In order to demonstrate the capability of actually deceiving a vein recognition algorithm, attack samples can be evaluated using a so called "2 scenario protocol", which is described in section \ref{subsec:2scenProt}. In the authors' previous works \cite{Schuiki21b, Schuiki21a}, an extensive threat evaluation was carried out for the finger vein databases introduced in \cite{PTome2015a, Qiu2017, Schuiki21a} by confronting twelve vein recognition schemes with the attack samples. The twelve recognition schemes can be categorized into three types of algorithms based on which features they extract from the vein images. Additionally, a presentation attack detection strategy was introduced that employs score level fusion of recognition algorithms. This was done by cross-evaluation of five fusion strategies along with three feature scaling approaches.

This study extends the authors' previous work by adding two feature extraction and comparison schemes that introduce a fourth algorithm category. Further, three additional feature normalization techniques and two additional fusion strategies are included in the experiments of this chapter. Besides that, this study transfers the threat analysis procedure to a private dorsal hand vein video database that was used in earlier studies \cite{Herzog20a, Schuiki20a}, but it was never shown whether it could actually deceive a vein recognition system.

The remainder of this chapter is structured as follows. Section \ref{sec:attack_dbs} describes the databases used later in this study. Section \ref{sec:threat} contains an evaluation of the threat that is emitted by the databases from section \ref{sec:attack_dbs} to 14 vein recognition schemes that can be categorized into 4 types of algorithms. In section \ref{sec:PAD}, the comparison scores from section \ref{sec:threat} are combined using score level fusion in order to carry out presentation attack detection. Finally, section \ref{sec:summary} includes a summary of this study.

\section{Attack Databases}
\label{sec:attack_dbs}

In total, four databases are included in this study which are described hereafter. 

\begin{enumerate}
    \item[A)] \textit{Paris Lodron University of Salzburg Palmar Finger Vein Spoofing Data Set (PLUS-FV3 Spoof)\footnote{https://wavelab.at/sources/PLUS-FV3-PALMAR-Image-Spoof/}:} The PLUS-FV3 Spoof data set uses a subset of the PLUS Vein-FV3 \cite{Kauba18c} database as bona fide samples. For the collection of presentation attack artefacts, binarized vein images from 6 fingers (i.e. index, middle and ring finger of both hands) of 22  subjects were printed on paper and sandwiched into a top and bottom made of beeswax. The binarization was accomplished by applying Principal Curvature \cite{BChoi09b} feature extraction in two different levels of vessel thickness, named \textit{thick} and \textit{thin}. The original database was captured with two types of light sources, namely \textit{LED} and \textit{Laser}. Therefore, presentation attacks were created for both illumination variants. While the original database was captured in 5 sessions per finger, only three of those were reused for presentation attack generation. Summarized, a total of 396 (22*6*3) presentation attacks per light source (LED \& Laser) and vein thickness (thick \& thin) with corresponding to 660 (22*6*5) bona fide samples are available. Every sample is of size 192x736.  
    \\
    \item[B)] \textit{The Idiap Research Institute VERA Fingervein Database (IDIAP VERA)\footnote{https://www.idiap.ch/en/dataset/vera-fingervein}:} The IDIAP VERA finger vein database consists of 440 bona fide images that correspond to 2 acquisition sessions of left and right hand index fingers of 110 subjects. Therefore these are considered as 220 unique fingers captured 2 times each. Every sample has one presentation attack counterpart. Presentation attacks are generated by printing preprocessed samples on high quality paper using a laser printer and enhancing vein contours with a black whiteboard marker afterwards. Every sample is provided in two modes named \textit{full} and \textit{cropped}. While the full set is comprised of the raw images captured with size 250x665, the cropped images were generated by removing a 50pixel margin from the border, resulting in images of size 150x565. 
    \\
    \item[C)] \textit{South China University of Technology Spoofing Finger Vein Database (SCUT-SFVD)\footnote{https://github.com/BIP-Lab/SCUT-SFVD}:} The SCUT-SFVD database was collected from 6 fingers (i.e. index, middle and ring finger of both hands) of 100 persons captured in 6 acquisition sessions, making a total of 3600 bona fide samples. For presentation attack generation, each finger vein image is printed on two overhead projector films which are aligned and stacked. In order to reduce overexposure, additionally a strong white paper (200$g/m^2$) is put in-between the two overhead projector films. Similar to the IDIAP VERA database, the SCUT-SFVD is provided in two modes named \textit{full} and \textit{roi}. While in the full set every image sample has a resolution of 640x288 pixel, the samples from the roi set are of variable size. Since the LBP and the ASAVE matching algorithm can not be evaluated on variable sized image samples, a third set was generated for this study named \textit{roi-resized} where all roi samples have been resized to 474x156 which corresponds to the median of all heights and widths from the roi set.
    \\ 
    \item[D)] \textit{Paris Lodron University of Salzburg Dorsal Hand Vein Video Spoofing Data Set (PLUS-DHV Spoof):} This database was initially created by \cite{Herzog20a} in order to analyze a finger vein video presentation attack detection scheme proposed by \cite{Raghavendra2015b}. The database consists of video samples from both hands of 13 participants, that were captured using two different illumination variants: \textit{Reflected light}, where the illumination source is placed next to the imaging sensor in order to capture the light which is not absorbed but reflected by the user's hand, as well as \textit{transillumination}, where the light source comes from the opposite side and goes through the user's hand. For every bona fide video sample, five different video attacks exist: (i) printed on paper using a laser printer (\textit{Paper Still}), (ii) printed on paper with applied movement back and forth (\textit{Paper Moving}), (iii) shown on a smartphone display (\textit{Display Still}), (iv) shown on smartphone display with programmed sinusoidal translation oscillation along the x axis (\textit{Display Moving}) and (v) shown on smartphone display with programmed sinusoidal scaling oscillation in every direction (\textit{Display Zoom}). For this study, 10 equally spaced frames were extracted throughout every video sequence, resulting in 260 (26*10) bona fide as well as attack samples, per attack type. Further, region of interest preprocessing was applied. Note however, that unfortunately this database is a private one.

\end{enumerate}

\section{Threat Analysis}
\label{sec:threat}

This section follows the idea from previous publications \cite{Schuiki21a, Schuiki21b}, in which databases A)\,-\,C) from section \ref{sec:attack_dbs} were subjected to experiments in order to evaluate the threat these attack samples emit to a variety of recognition algorithms. The goal of the experiments in this section is (i) to carry out the experiments from previous publications with a slightly different setting in the options as well as (ii) the transfer of this threat evaluation to the dorsal hand vein database D) explained in section \ref{sec:attack_dbs}. To do so, the threat evaluation protocol described in section \ref{subsec:2scenProt} is used for the experiments in this section. Altogether 14 distinct feature extraction and comparison schemes are used in this study that can be categorized into four classes of algorithms based on the type of feature they extract from the vein samples: 

\begin{itemize}
    \item \textit{Binarized vessel networks}: Algorithms from this category work by transforming a raw vein image into a binary image where the background (and also other parts of the human body such as flesh) is removed and only the extracted vessel structures remain. The binarized image is then used as a feature image for the comparison step. Seven different approaches are included in this study that finally create such a binarized vein image. \textit{Maximum Curvature (MC) \cite{Miura07a}} and  \textit{Repeated Line Tracking (RLT) \cite{Miura2004a}} try to achieve this by looking at the cross sectional profile of the finger vein image. Other methods such as \textit{Wide Line Detector (WLD) \cite{BHuang2010b}}, \textit{Gabor Filter (GF) \cite{AKumar2012a}} and \textit{Isotropic Undecimated Wavelet Transform (IUWT) \cite{JStarck2007a}} also consider local neighbourhood regions by using filter convolution. A slightly different approach is given by \textit{Principal Curvature (PC) \cite{BChoi09b}} which first computes the normalized gradient field and then looks at the eigenvalues of the Hessian matrix at each pixel. All so far described binary image extraction methods use a correlation measure to compare probe and template samples which is often referred to as \textit{Miura-matching} due to its introduction in Miura et al. \cite{Miura2004a}. One more sophisticated vein pattern based feature extraction and matching strategy is \textit{Anatomy Structure Analysis-Based Vein Extraction (ASAVE) \cite{LYang2018a}}, which includes two different techniques for binary vessel structure extraction as well as a custom matching strategy. 
    
    \item \textit{Keypoints}: The term keypoint is generally understood as a specific pixel or pixel region in an digital image that provides some interesting information to a given application. Every keypoint is stored by describing its local neighbourhood and its location. This research uses three keypoint based feature extraction and matching schemes. One such keypoint detection method, known as \textit{Deformation Tolerant Feature Point Matching (DTFPM) \cite{Matsuda2016a}}, was especially tailored for the task of finger vein recognition. This is achieved by considering shapes that are common in finger vein structures. Additionally, modified versions of general purpose keypoint detection and matching schemes, \textit{SIFT} and \textit{SURF}, as described in \cite{Kauba14a} are tested in this research. The modification includes filtering such that only keypoints inside the finger are used while keypoints at the finger contours or even in the background are discarded.
    
    \item \textit{Texture information}: Image texture is a feature that describes the structure of an image. Shapiro and Stockman \cite{Shapiro2001} define image texture as something that gives information about the spatial arrangement of color or intensities in an image or selected region of an image. While two images can be identical in terms of their histograms, they can be very different when looking at their spatial arrangement of bright and dark pixels. Two methods are included in this work that can be counted to texture-based approaches. One of which is a \textit{Local Binary Pattern \cite{ECLee2009a}} descriptor that uses histogram intersection as a similarity metric. The second method is a \textit{convolutional neural network (CNN)} based approach that uses triplet loss as presented in \cite{Wimmer20a}. Similarity scores for the CNN approach are obtained by computing the inverse Euclidean distance given two feature vectors corresponding to two finger vein samples. 
    
    \item \textit{Minutiae-based}: The term minutiae descends from the domain of fingerprint biometrics. Every fingerprint is a unique pattern that consists of ridges and valleys. The locations where such a pattern has discontinuities such as ridge endings or bifurcations are named "minutiae points". This concept of finding such minutiae points was successfully transferred \cite{Linortner21a} to the vein biometrics domain by skeletonization of an extracted binarized vein image, such as described in the first category. This study now employs two schemes in order to perform vein recognition, that both use these extracted minutiae points. First, the proprietary software \textit{VeriFinger SDK (VF)}     \footnote{https://www.neurotechnology.com/verifinger.html} is used for comparison of these minutiae points. A second method, \textit{Location-based Spectral Minutiae Representation (SML)} \cite{Xu2009} uses the minutiae points as an input in order to generate a representation that can finally be compared using a correlation measure.   

\end{itemize}

\subsection{Threat Evaluation Protocol}
\label{subsec:2scenProt}

To evaluate the level of threat exhibited by a certain database, a common evaluation scheme is used that employs two consecutive steps, hence often coined “2 scenario protocol” \cite{Chingovska2019, Patil2016, PTome2014a}, is adopted in this study. The two scenarios are briefly summarized hereafter (description taken from \cite{Schuiki21b}):

\begin{itemize}
    \item \textbf{Licit Scenario (Normal Mode)}: The first scenario employs two types of users: Genuine (positives) and zero effort impostors (negatives). Therefore, both enrollment and verification is accomplished using bona fide finger vein samples. Through varying the decision threshold, the False Match Rate (FMR, \i.e. the ratio of wrongly accepted impostor attempts to the number of total impostor attempts) and the False Non Match Rate (FNMR, \i.e. the ratio of wrongly denied genuine attempts to the total number of genuine verification attempts) can be determined. The normal mode can be understood as a matching experiment which has the goal to determine an operating point for the second scenario. The operating point is set at the threshold value where the FMR = FNMR (i.e. Equal Error Rate).
    
    \item \textbf{Spoof Scenario (Attack Mode)}: The second scenario uses genuine (positives) and presentation attack (negatives) users.
    Similar to the first scenario, enrollment is accomplished using bona fide samples. Verification attempts are performed by matching presentation attack samples against their corresponding genuine enrollment samples or templates. Given the threshold from the licit scenario, the proportion of wrongly accepted presentation attacks is then reported as the Impostor Attack Presentation Match Rate (IAPMR), as defined by the ISO/IEC 30107-3:2017 \cite{iso_30107}.

\end{itemize}

\subsection{Experimental Results}

\begin{figure*}[ht]
\includegraphics[width=\textwidth]{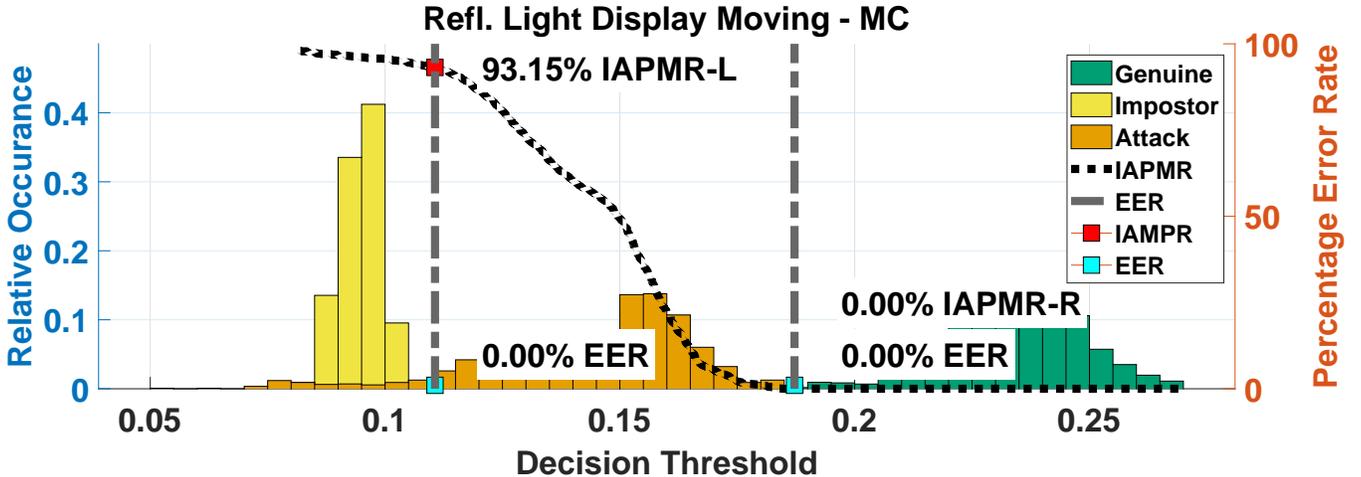}
\centering
\caption{Exemplary visualization of the outcomes from the two scenario protocol when the Maximum Curvature scheme was confronted with hand vein attack samples from type: reflected light display moving.}
\label{fig:2Scen_prot_DHV}
\end{figure*}

The experimental results are divided into two tables: Table \ref{tab:ThreatAnalysis_fv} contains the outcomes from the threat evaluation on the finger vein databases (PLUS, IDIAP Vera \& SCUT) and table \ref{tab:ThreatAnalysis_dhv} contains the results from the experiments on the dorsal hand vein database. Note that for databases where multiple types of attacks exist for the same corresponding bona fide samples, the reported EER values are equal since the error rate calculation is solely based on bona fide data. The table that contains the threat analysis for the hand vein samples (table \ref{tab:ThreatAnalysis_dhv}) includes always two IAPMR values per attack type. The reason for this is illustrated in figure \ref{fig:2Scen_prot_DHV}: Due to the fact that only one video sequence is available per subject to extract frames from, the samples are somewhat biased in terms of intra-class variability. Hence, genuine and impostor scores can often be separated perfectly such that a whole range of decision thresholds would be eligible options to set the EER (the range is illustrated in the figure as the region in-between the vertical dash-dotted lines). The dash-dotted lines represent the extreme cases where the EER would still be zero, i.e. perfect separation of genuine and impostor scores. When now looking at the two IAPMRs at the extreme cases, one can see that this rate varies (in this case) between 0.00\% and 93.15\%. Thus, the IAPMRs for the hand vein database is always reported as for the left-most (IAPMR-L) and right-most (IAPMR-R) extreme case. Note that for EERs $\neq\,0$ both IAPMRs are identical.

The minutiae based recognition schemes are evaluated using the \textit{VeinPLUS+ Framework} \cite{Linortner21b}. The binary vessel network based methods, the keypoint based methods and the LBP scheme were evaluated using the \textit{PLUS OpenVein Toolkit} \cite{Kauba19a}. The CNN based approach was implemented in Python. Due to the fact that this is a learning based approach and such learning is non-deterministic, the EERs and IAPMRs for the CNN method are calculated by taking the arithmetic mean over a 2-fold-cross validation. This is done for every database except for the PLUS LED and PLUS Laser, since there exists a second open source database \cite{Sequeira18a} which descends from a similar imaging sensor that is used for training the CNN. For comparison the 'FVC' protocol, named after the Fingerprint Verification Contest 2004 \cite{Maio2004} where it was introduced, was used. This protocol executes every possible genuine comparison but only compares the first sample of each subject with the first sample of all remaining subjects. Hence, it is ensured that every genuine comparison is made, but largely reduces the amount of impostor scores.


\begin{table*}[ht]
\caption{Vulnerability analysis using 2 scenario protocol for the finger vein DBs.} 
\centering 
\resizebox{\textwidth}{!}{
\begin{tabular}{c cccccccccccccc} 

\toprule
 & MC & PC & GF & IUWT & RLT & WLD & ASAVE & DTFPM & SIFT & SURF & LBP & CNN & SML & VF \\
\multicolumn{15}{l}{\textbf{PLUS LED Thick}} \rule{0pt}{4ex} \\
\midrule

EER & 0.60 & 0.75 & 1.06 & 0.68 & 4.62 & 1.14 & 2.35 & 2.35 & 1.06 & 3.33 & 3.86 & 2.90 & 2.79 & 0.80\\
IAPMR & 72.81 & 69.93 & 36.99 & 79.08 & 40.65 & 69.15 & 24.31 & 16.34 & 0.00 & 0.00 & 0.00 & 0.67 & 4.02 & 5.16\\

\multicolumn{15}{l}{\textbf{PLUS LED Thin}} \rule{0pt}{4ex}\\
\midrule

EER & 0.60 & 0.75 & 1.06 & 0.68 & 4.62 & 1.14 & 2.35 & 2.35 & 1.06 & 3.33 & 3.86 & 2.90 & 2.79 & 0.80\\
IAPMR & 89.65 & 79.80 & 59.72 & 89.77 & 32.83 & 83.96 & 19.07 & 15.78 & 0.00 & 0.00 & 0.25 & 0.35 & 5.87 & 5.24\\

\multicolumn{15}{l}{\textbf{PLUS Laser Thick}} \rule{0pt}{4ex}\\
\midrule

EER & 1.29 & 1.65 & 2.72 & 1.97 & 6.21 & 2.80 & 2.50 & 2.90 & 1.04 & 3.71 & 4.27 & 6.82 & 2.95 & 1.01\\
IAPMR & 59.90 & 57.47 & 31.55 & 79.44 & 22.48 & 57.22 & 9.07 & 4.98 & 0.00 & 0.00 & 0.00 & 0.00 & 2.67 & 6.54\\

\multicolumn{15}{l}{\textbf{PLUS Laser Thin}} \rule{0pt}{4ex}\\
\midrule

EER & 1.29 & 1.65 & 2.72 & 1.97 & 6.21 & 2.80 & 2.50 & 2.90 & 1.04 & 3.71 & 4.27 & 6.82 & 2.95 & 1.01\\
IAPMR & 76.52 & 66.92 & 52.65 & 84.34 & 16.79 & 77.78 & 2.02 & 5.18 & 0.13 & 0.00 & 0.00 & 0.05 & 3.60 & 5.62\\

\multicolumn{15}{l}{\textbf{IDIAP Vera FV Full}} \rule{0pt}{4ex}\\
\midrule

EER & 2.66 & 2.73 & 6.83 & 4.95 & 29.55 & 6.03 & 9.11 & 10.45 & 4.54 & 11.39 & 8.17 & 6.35 & 6.69 & 4.04\\
IAPMR & 93.18 & 90.45 & 85.76 & 93.18 & 35.00 & 93.48 & 72.58 & 26.21 & 14.24 & 0.91 & 16.97 & 17.27 & 59.09 & 21.67\\

\multicolumn{15}{l}{\textbf{IDIAP Vera FV Cropped}} \rule{0pt}{4ex}\\
\midrule

EER & 5.52 & 6.91 & 11.36 & 6.36 & 27.19 & 9.09 & 19.10 & 6.69 & 5.43 & 11.62 & 5.80 & 9.93 & 17.27 & 11.11\\
IAPMR & 92.27 & 89.24 & 81.97 & 91.82 & 38.64 & 90.45 & 68.79 & 81.97 & 44.55 & 14.24 & 69.09 & 7.84 & 64.55 & 21.97\\

\multicolumn{15}{l}{\textbf{SCUT-SFVD Full}} \rule{0pt}{4ex}\\
\midrule

EER & 4.01 & 4.79 & 9.40 & 6.29 & 14.01 & 7.60 & 11.56 & 8.90 & 2.37 & 5.49 & 8.78 & 0.74 & 7.63 & 1.41\\
IAPMR & 86.33 & 84.67 & 54.90 & 74.06 & 40.36 & 74.21 & 74.98 & 73.75 & 30.43 & 4.18 & 45.43 & 68.65 & 59.76 & 32.90\\

\multicolumn{15}{l}{\textbf{SCUT-SFVD ROI-Resized}} \rule{0pt}{4ex}\\
\midrule

EER & 2.28 & 2.12 & 3.52 & 2.10 & 2.94 & 2.36 & 6.03 & 5.63 & 1.92 & 9.41 & 3.51 & 0.84 & 7.25 & 3.78\\
IAPMR & 60.36 & 47.95 & 36.88 & 49.85 & 9.46 & 52.49 & 59.88 & 55.08 & 34.93 & 7.42 & 55.36 & 55.00 & 61.22 & 40.40\\

\bottomrule
\end{tabular}
}
\label{tab:ThreatAnalysis_fv}
\end{table*}

\subsubsection{Finger Vein Attack Databases}

When confronted with the finger vein attacks, binary vessel pattern based schemes tend to produce IAPMRs up to 93.48\% for the IDIAP Vera, 86.33\% for the SCUT and 89.54\% for the PLUS databases. Such IAPMRs indicate that roughly up to 9 out of 10 attack samples could wrongly be classified as a bona fide presentation. Compared to the reference paper \cite{Schuiki21b}, the "Miura matching" comparison step for the region of interest (roi/cropped) versions was also executed with parameters that allow for a certain translation invariance. Doing so, some EER results could be reduced by up to 15\%. As a consequence also the IAPMRs increased, giving a clearer image on the threat that these cropped samples exhibit to the algorithms that extract a binary feature image.
General purpose keypoint based recognition schemes, especially SURF, seem relatively unimpressed by the attack samples, reaching an overall high IAPMR of 14.24\% at the IDIAP Vera cropped attacks. The keypoint method that was especially tailored for the vein recognition task, DTFPM, along with the texture based and minutiae based methods show a relatively in-homogeneous behaviour. All, however, indicate that the PLUS attacks are little to no threat to them compared to the IDIAP and SCUT attack samples.


\begin{table*}[ht]
\caption{Results of the threat analysis for the hand vein attacks.}
\centering 
\resizebox{\textwidth}{!}{
\begin{tabular}{c cccccccccccccc} 
\toprule

 & MC & PC & GF & IUWT & RLT & WLD & ASAVE & DTFPM & SIFT & SURF & LBP & CNN & SML & VF \\

\multicolumn{15}{l}{\textbf{PLUS-DHV Transillumination Paper-Still}} \rule{0pt}{4ex}\\
\midrule

EER & 0.00 & 0.00 & 0.00 & 0.00 & 2.66 & 0.00 & 0.00 & 0.00 & 0.00 & 0.00 & 0.00 & 1.62 & 0.17 & 0.09\\
IAPMR-R & 0.00 & 0.00 & 0.00 & 0.00 & 0.00 & 0.00 & 0.00 & 0.00 & 0.00 & 0.00 & 0.00 & 3.31 & 3.15 & 2.31\\
IAPMR-L & 94.90 & 5.03 & 0.28 & 6.01 & 0.00 & 11.75 & 38.95 & 85.52 & 0.35 & 0.00 & 6.78 & 3.31 & 3.15 & 2.31\\

\multicolumn{15}{l}{\textbf{PLUS-DHV Transillumination Paper-Moving}} \rule{0pt}{4ex}\\
\midrule

IAPMR-R & 0.00 & 0.00 & 0.00 & 0.00 & 0.00 & 0.00 & 0.00 & 0.00 & 0.00 & 0.00 & 0.00 & 3.38 & 3.78 & 2.24\\
IAPMR-L & 92.38 & 4.97 & 0.42 & 5.87 & 0.00 & 10.77 & 32.94 & 79.86 & 0.00 & 0.00 & 3.85 & 3.38 & 3.78 & 2.24\\ 

\multicolumn{15}{l}{\textbf{PLUS-DHV Transillumination Display-Still}} \rule{0pt}{4ex}\\
\midrule

IAPMR-R & 1.40 & 0.49 & 24.20 & 0.56 & 8.46 & 10.77 & 0.00 & 0.00 & 0.00 & 0.00 & 0.00 & 18.58 & 5.31 & 9.51\\
IAPMR-L & 99.65 & 33.15 & 32.87 & 42.24 & 8.46 & 61.82 & 2.17 & 96.22 & 3.64 & 0.00 & 0.00 & 18.46 & 5.31 & 9.51\\

\multicolumn{15}{l}{\textbf{PLUS-DHV Transillumination Display-Moving}} \rule{0pt}{4ex}\\
\midrule

IAPMR-R & 0.63 & 0.28 & 23.01 & 1.19 & 7.13 & 10.84 & 0.00 & 0.00 & 0.00 & 0.00 & 0.00 & 18.54 & 5.59 & 10.77\\
IAPMR-L & 98.95 & 34.83 & 29.93 & 40.70 & 7.13 & 54.69 & 5.66 & 95.03 & 3.43 & 0.00 & 0.00 & 18.50 & 5.59 & 10.77\\
 
\multicolumn{15}{l}{\textbf{PLUS-DHV Transillumination Display-Zoom}} \rule{0pt}{4ex}\\
\midrule

IAPMR-R & 0.14 & 0.21 & 22.87 & 0.21 & 9.02 & 10.56 & 0.00 & 0.00 & 0.00 & 0.00 & 0.00 & 18.54 & 6.08 & 9.72\\
IAPMR-L & 99.65 & 33.22 & 31.47 & 43.29 & 9.02 & 57.55 & 2.87 & 96.50 & 3.43 & 0.00 & 0.00 & 18.54 & 6.08 & 9.72\\

\multicolumn{15}{l}{\textbf{PLUS-DHV Reflected Light Paper-Still}} \rule{0pt}{4ex}\\
\midrule

EER & 0.00 & 0.00 & 0.10 & 0.00 & 3.25 & 0.00 & 0.00 & 0.00 & 0.00 & 0.18 & 0.00 & 2.62 & 0.34 & 0.69\\
IAPMR-R & 0.00 & 0.00 & 2.66 & 0.00 & 0.00 & 13.50 & 0.00 & 0.00 & 0.00 & 0.00 & 0.00 & 0.00 & 1.47 & 0.42\\
IAPMR-L & 92.80 & 0.00 & 2.66 & 0.07 & 0.00 & 13.50 & 17.90 & 8.18 & 0.00 & 0.00 & 0.00 & 0.00 & 1.47 & 0.42\\
 
\multicolumn{15}{l}{\textbf{PLUS-DHV Reflected Light Paper-Moving}} \rule{0pt}{4ex}\\
\midrule

IAPMR-R & 0.00 & 0.00 & 2.52 & 0.00 & 0.00 & 10.56 & 0.00 & 0.00 & 0.00 & 0.00 & 0.00 & 0.00 & 1.75 & 0.42\\
IAPMR-L & 91.96 & 0.00 & 2.52 & 0.63 & 0.00 & 10.56 & 19.65 & 23.43 & 0.00 & 0.00 & 0.00 & 0.00 & 1.75 & 0.42\\

\multicolumn{15}{l}{\textbf{PLUS-DHV Reflected Light Display-Still}} \rule{0pt}{4ex}\\
\midrule

IAPMR-R & 0.00 & 8.46 & 12.24 & 2.45 & 64.83 & 0.00 & 0.00 & 0.00 & 0.07 & 3.08 & 0.00 & 8.00 & 1.33 & 3.08\\
IAPMR-L & 94.97 & 67.90 & 12.24 & 30.07 & 64.83 & 0.00 & 13.15 & 54.62 & 3.57 & 3.08 & 0.00 & 8.00 & 1.33 & 3.08\\

\multicolumn{15}{l}{\textbf{PLUS-DHV Reflected Light Display-Moving}} \rule{0pt}{4ex}\\
\midrule

IAPMR-R & 0.00 & 7.97 & 12.38 & 2.59 & 65.38 & 0.00 & 0.00 & 0.00 & 0.00 & 2.45 & 0.00 & 8.54 & 1.75 & 3.29\\
IAPMR-L & 93.15 & 71.33 & 12.38 & 37.34 & 65.38 & 0.00 & 19.86 & 64.13 & 3.43 & 2.45 & 0.00 & 8.54 & 1.75 & 3.29\\
 
\multicolumn{15}{l}{\textbf{PLUS-DHV Reflected Light Display-Zoom}} \rule{0pt}{4ex}\\
\midrule

IAPMR-R & 0.49 & 9.58 & 11.96 & 2.73 & 59.16 & 0.00 & 0.00 & 0.00 & 0.00 & 2.38 & 0.00 & 6.42 & 1.54 & 3.15\\
IAPMR-L & 95.31 & 73.43 & 11.96 & 36.92 & 59.16 & 0.00 & 12.66 & 58.39 & 3.36 & 2.38 & 0.00 & 6.42 & 1.54 & 3.15\\

\bottomrule
\end{tabular}
}
\label{tab:ThreatAnalysis_dhv}
\end{table*}

\subsubsection{Dorsal Hand Vein Attack Database}

When looking at the threat analysis for the dorsal hand vein attacks (table \ref{tab:ThreatAnalysis_dhv}), one is able to observe that (i) attack samples that were captured through transillumination are more likely to pose a threat to the recognition schemes and also (ii) attacks shown on a smartphone display tend to work better than the laser printed paper attacks. One exception constitutes the MC recognition scheme. It seems that, when treating both extreme cases IAPMR-R and IAPMR-L as equally valid, all attacks have an overall high potential to fool the MC algorithm. The same holds true for the DTFPM keypoint recognition scheme. For other binary vessel network schemes, as mentioned, only the display attacks seem to have a certain attack potential. The remaining general purpose keypoint schemes, SIFT and SURF, the Minutiae based schemes as well as LBP seem to remain unaffected regardless of the attack type used. Interestingly, the display attacks seem to have at least some attack potential to the CNN recognition scheme. 




\section{Attack Detection Using Score Level Fusion}
\label{sec:PAD}

Using the the observation that different recognition schemes vary in their behaviour and thus probably contain complementary information, this section includes experiments to combine multiple recognition schemes in order to achieve presentation attack detection. Note that throughout the following section, genuine scores (i.e. scores that descend from intra-subject comparisons) are viewed as bona fide scores and scores that descend from comparisons where an attack sample is compared to its bona fide counterpart are considered to be presentation attack scores. Solving the presentation attack problem via combination of multiple similarity scores descending from distinct recognition schemes can be formally written as follows. Let $s_{ij}$ be the $i$th comparison score out of $m$ total eligible comparisons that was assigned by the $j$th recognition scheme out of $n$ considered recognition schemes at a time, i.e. $i \in \{1,...,m\}$ and $j \in \{1,...,n\}$. Note that for an $i$th comparison, every recognition scheme needs to compare the same biometric samples. Further, let $x_i$ be the the vector $x_i\,=\,\left(s_{i1}, ..., s_{in}\right)$ that describes the $i$th comparison by concatenation of all $n$ recognition schemes considered at a time. Let $\mathrm{\bf{X}}$ be the set of all $m$ eligible comparisons from a certain database $\mathrm{\bf{X}} = \{x_1, ..., x_m\}$. Since some of the following fusion strategies demand for training data, the set of all comparisons $\mathrm{\bf{X}}$ is always divided into a train $\mathrm{\bf{X_{train}}}$ and a test split $\mathrm{\bf{X_{test}}}$ using k-fold cross validation. The combined score that remains after score level fusion for the $i$th comparison shall be denoted as $S_i$. The comparison can then be classified as bona fide or attack comparison by simple thresholding of $S_i$. 
\\

For the experiments in this study, three simple fusion strategies (\textit{Min-Rule Fusion}, \textit{Max-Rule Fusion} and \textit{Simple Sum-Rule Fusion}) were adopted from \cite{Snelick2003a} that can be seen in the following equations.

\begin{itemize}
    \item \textit{Min-Rule Fusion}
    \begin{equation} S_i = min( x_i ) \end{equation}
    \item \textit{Max-Rule Fusion}
    \begin{equation} S_i = max( x_i ) \end{equation}
    \item \textit{Simple Sum-Rule Fusion}
    \begin{equation} S_i = \sum_{j=1}^{n} s_{ij} \end{equation}
\end{itemize}

A slightly more sophisticated method is given by the \textit{Weighted Sum-Rule Fusion}. In \cite{Snelick2005} this technique is also named as "Matcher Weighting" in order to distinguish the weighting strategy from the one where every user or individual would get different weights instead of every recognition scheme.

\begin{itemize}
    \item \textit{Weighted Sum-Rule Fusion}
    \begin{equation} 
        S_i = \sum_{j=1}^{n} s_{ij} * w_j 
    \label{eq:wsum_rule_fusion}
    \end{equation}
\end{itemize}

In order to use the weighted sum rule fusion, a strategy has to be chosen on how to assign the weights $w_j$ in equation \ref{eq:wsum_rule_fusion}. Snelick et al. \cite{Snelick2005} propose to choose weights indirect proportional to the error rates. Since it is not intuitively clear which error rates to use in this study (since the goal is to use score level fusion of different recognition schemes to achieve presentation attack detection), the training split $\mathrm{\bf{X_{train}}}$ is used to calculate an Equal Error Rate estimate $EER_j$ per recognition scheme $j$. This decision can be justified since $\mathrm{\bf{X_{train}}}$ consists of bona fide and attack comparisons and therefore gives a rough estimate on how much a particular recognition scheme contributes to the PAD-task.

\begin{equation}
    w_j = \frac{\frac{1}{EER_j}}{\sum_{v=1}^{n} \frac{1}{EER_v} }
\end{equation}

Additionally, three binary classifiers are evaluated that are trained using $\mathrm{\bf{X_{train}}}$ in order to predict the test data $\mathrm{\bf{X_{test}}}$ per fold: (i) \textit{Fisher Linear Discriminant}, (ii) \textit{Support Vector Machine with Linear Kernel} and (iii) \textit{Support Vector Machine with Radial Basis Function Kernel}.
\\

Similarity scores from distinct recognition algorithms, however, do not necessary lie in the same range. Therefore a variety of comparison score normalization strategies are applied to $x_i$ in order to create a normalized comparison feature vector $x_i'$ along with the option of applying no feature scaling at all (\textit{No Norm:} $x_i' = x_i$). Note that calculations over the whole data (e.g. mean, min, max,...) are also conducted over the training split $\mathrm{\bf{X_{train}}}$ for the normalization step as well in order to simulate a realistic scenario where the sample under test is not included in parameter determination. Together with the option of omitting score normalization, six strategies are included in this research. 

Three popular score normalization techniques \cite{Jain2005a}, \textit{Min-Max Norm}, \textit{Z-Score Norm} and \textit{Tanh-Norm} are described in the following equations, where $\sigma$ represents the operator for calculation of the standard deviation and $\mu$ stands for calculation of the arithmetic mean. 

\begin{itemize}
    \item \textit{Min-Max Norm}
    \begin{equation} x_i' = \frac{x_i - min(\mathrm{\bf{X_{train}}})}{max(\mathrm{\bf{X_{train}}}) - min(\mathrm{\bf{X_{train}}})} \end{equation}
    \item \textit{Tanh-Norm}
    \begin{equation} x_i' = 0.5 * \left( tanh \left( 0.01 * \frac{x_i - \mu(\mathrm{\bf{X_{train}}})}{\sigma(\mathrm{\bf{X_{train}}})} \right) + 1 \right) \end{equation}
    \item \textit{Z-Score Norm}
    \begin{equation} x_i' = \frac{x_i - \mu(\mathrm{\bf{X_{train}}})}{\sigma(\mathrm{\bf{X_{train}}})} \end{equation}
\end{itemize}

Another normalization technique was proposed by He et al. \cite{He2010} named \textit{Reduction of high-scores effect (RHE) normalization}. Here, $\mathrm{\bf{X_{train_{bf}}}}$ indicates to use only the bona fide comparison scores for calculation of the mean and standard deviation. 

\begin{itemize}
    \item \textit{Rhe-Norm}
    \begin{equation} x_i' = \frac{x_i - min(\mathrm{\bf{X_{train}}})}{\mu(\mathrm{\bf{X_{train_{bf}}}}) + \sigma(\mathrm{\bf{X_{train_{bf}}}}) - min(\mathrm{\bf{X_{train}}})} \end{equation}
\end{itemize}

Additionally, rescaling the feature vector $x$ to unit length is tested. Note that for the cases where only a single recognition scheme is considered, the \textit{Unit Length Norm} would result in ones ( i.e. $\frac{x}{||x||} = 1$ for the case where x is a scalar). Therefore the normalization is omitted for these cases.

\begin{itemize}
    \item \textit{Unit Length Norm}
    \begin{equation} x_i' = \frac{x_i}{||x_i||} \end{equation}
\end{itemize}

The ISO/IEC 30107-3:2017 \cite{iso_30107} defines decision threshold dependent metrics for presentation attack detection such as Attack Presentation Classification Error Rate (APCER) and Bona Fide Presentation Classification Error Rate (BPCER):

\begin{itemize}
    \item \textit{ Attack Presentation Classification Error Rate (APCER)}: Proportion of attack presentations incorrectly classified as bona fide presentations in a specific scenario
    
    
    \item \textit{ Bona Fide Presentation Classification Error Rate (BPCER)}: Proportion of bona fide presentations incorrectly classified as presentation attacks in a specific scenario
    

\end{itemize}

The PAD performance is reported in table \ref{tbl:PAD} in terms of detection equal error rate D-EER (operating point where BPCER = APCER), BPCER20 (BPCER at APCER $<=$ 0.05) and BPCER100 (BPCER at APCER $<=$ 0.01).

\subsection{Experimental Results}

For the score level fusion experiments described above, results are obtained by conducting an exhaustive cross combination that includes all of the aforementioned fusion and normalization strategies. Per database, $2^{14}-1 = 16\,383$ method constellations (since 14 different recognition schemes are used) exist, all of which are further evaluated using 6*7=42 Norm-Fusion combinations. Doing so allows for a broader perspective on which score level fusion techniques tend to work well very often. The dorsal hand vein database was not included in these experiments since for every attack case there is at least one recognition scheme that achieves 0.00\% IAPMR, meaning that at least one single method would suffice to separate bona fide from attack samples, thus annulling the reason of performing score level fusion.

\begin{table*}[ht]
\caption{ Selection of best working method constellations in terms of detection error rate. Note that for the PLUS databases, there are often multiple eligible options. This is only one constellation that uses as little recognition schemes as possible.}
\centering
\resizebox{\textwidth}{!}{
\begin{tabular}{ccccccccccccccccccc cc}

\toprule

Database & \rot{D-EER} & \rot{BPCER20} & \rot{BPCER100}
& \rot{Fusion} & \rot{Norm}      
& \rot{MC} & \rot{PC} & \rot{WLD} & \rot{RLT} & \rot{GF} & \rot{IUWT} & \rot{ASAVE} & \rot{DTFPM} & \rot{SURF} & \rot{SIFT} & \rot{LBP} & \rot{CNN} & \rot{SML} & \rot{VF} \\                                

\midrule

PLUS-LED& \multirow{2}{*}{0.00} & \multirow{2}{*}{0.00} & \multirow{2}{*}{0.00} & \multirow{2}{*}{svm-rbf} & \multirow{2}{*}{z-norm} & \multirow{2}{*}{}& \multirow{2}{*}{}& \multirow{2}{*}{}& \multirow{2}{*}{}& \multirow{2}{*}{}& \multirow{2}{*}{}& \multirow{2}{*}{}& \multirow{2}{*}{}& \multirow{2}{*}{}& \multirow{2}{*}{\OK} & \multirow{2}{*}{}& \multirow{2}{*}{\OK} & \multirow{2}{*}{} & \multirow{2}{*}{}  \\ thick \\

PLUS-LED& \multirow{2}{*}{0.00} & \multirow{2}{*}{0.00} & \multirow{2}{*}{0.00} & \multirow{2}{*}{sum-rule} & \multirow{2}{*}{z-norm} & \multirow{2}{*}{}& \multirow{2}{*}{}& \multirow{2}{*}{}& \multirow{2}{*}{}& \multirow{2}{*}{}& \multirow{2}{*}{}& \multirow{2}{*}{}& \multirow{2}{*}{}& \multirow{2}{*}{\OK} & \multirow{2}{*}{\OK} & \multirow{2}{*}{} & \multirow{2}{*}{} & \multirow{2}{*}{} & \multirow{2}{*}{} \\thin \\

PLUS-Laser& \multirow{2}{*}{0.00} & \multirow{2}{*}{0.00} & \multirow{2}{*}{0.00} & \multirow{2}{*}{sum-rule} & \multirow{2}{*}{no-norm} & \multirow{2}{*}{}& \multirow{2}{*}{}& \multirow{2}{*}{}& \multirow{2}{*}{}& \multirow{2}{*}{}& \multirow{2}{*}{}& \multirow{2}{*}{}& \multirow{2}{*}{}& \multirow{2}{*}{}& \multirow{2}{*}{\OK} & \multirow{2}{*}{}& \multirow{2}{*}{\OK}& \multirow{2}{*}{}& \multirow{2}{*}{}  \\thick \\

PLUS-Laser& \multirow{2}{*}{0.00} & \multirow{2}{*}{0.00} & \multirow{2}{*}{0.00} & \multirow{2}{*}{svm-lin} & \multirow{2}{*}{z-norm} & \multirow{2}{*}{}& \multirow{2}{*}{} & \multirow{2}{*}{}& \multirow{2}{*}{}& \multirow{2}{*}{}& \multirow{2}{*}{}& \multirow{2}{*}{}& \multirow{2}{*}{} & \multirow{2}{*}{}& \multirow{2}{*}{\OK} & \multirow{2}{*}{\OK}& \multirow{2}{*}{\OK}  & \multirow{2}{*}{} & \multirow{2}{*}{} \\thin \\

IDIAP VERA & \multirow{2}{*}{1.14} & \multirow{2}{*}{0.91} & \multirow{2}{*}{1.82} & \multirow{2}{*}{fisher-lda} & \multirow{2}{*}{unit-length} & \multirow{2}{*}{}& \multirow{2}{*}{\OK}& \multirow{2}{*}{\OK} & \multirow{2}{*}{\OK}& \multirow{2}{*}{\OK}& \multirow{2}{*}{}& \multirow{2}{*}{\OK} & \multirow{2}{*}{} & \multirow{2}{*}{\OK} & \multirow{2}{*}{\OK} & \multirow{2}{*}{\OK} & \multirow{2}{*}{\OK}  & \multirow{2}{*}{} & \multirow{2}{*}{} \\full \\

IDIAP VERA & \multirow{2}{*}{3.18} & \multirow{2}{*}{2.27} & \multirow{2}{*}{13.18} & \multirow{2}{*}{svm-lin} & \multirow{2}{*}{rhe-norm} & \multirow{2}{*}{\OK}& \multirow{2}{*}{\OK} & \multirow{2}{*}{\OK} & \multirow{2}{*}{\OK}& \multirow{2}{*}{\OK} & \multirow{2}{*}{\OK}& \multirow{2}{*}{\OK} & \multirow{2}{*}{\OK} & \multirow{2}{*}{\OK} & \multirow{2}{*}{\OK} & \multirow{2}{*}{}& \multirow{2}{*}{\OK} & \multirow{2}{*}{\OK} & \multirow{2}{*}{\OK}  \\cropped \\

SCUT-SFVD& \multirow{2}{*}{0.33} & \multirow{2}{*}{0.00} & \multirow{2}{*}{0.17} & \multirow{2}{*}{svm-lin} & \multirow{2}{*}{min-max} & \multirow{2}{*}{} & \multirow{2}{*}{} & \multirow{2}{*}{} & \multirow{2}{*}{\OK} & \multirow{2}{*}{} & \multirow{2}{*}{\OK} & \multirow{2}{*}{\OK} & \multirow{2}{*}{\OK} & \multirow{2}{*}{\OK} & \multirow{2}{*}{} & \multirow{2}{*}{\OK} & \multirow{2}{*}{\OK} & \multirow{2}{*}{} & \multirow{2}{*}{}  \\full \\

SCUT-SFVD & \multirow{2}{*}{0.76} & \multirow{2}{*}{0.67} & \multirow{2}{*}{0.67} & \multirow{2}{*}{svm-rbf} & \multirow{2}{*}{rhe-norm} & \multirow{2}{*}{} & \multirow{2}{*}{\OK} & \multirow{2}{*}{\OK} & \multirow{2}{*}{\OK} & \multirow{2}{*}{\OK}& \multirow{2}{*}{}& \multirow{2}{*}{\OK} & \multirow{2}{*}{} & \multirow{2}{*}{\OK} & \multirow{2}{*}{} & \multirow{2}{*}{\OK} & \multirow{2}{*}{\OK} & \multirow{2}{*}{} & \multirow{2}{*}{\OK}  \\roi-resized \\

\bottomrule

\end{tabular}
}
\label{tbl:PAD}
\end{table*}

\begin{figure}
\caption{Boxplots including cross combinations of normalization- and fusion-strategies. Note that every boxplot contains 16\,383 values for $2^{14}-1$ possible method constellations. To see what number on the x axis corresponds to which norm-fusion-combination, table \ref{tab:boxplot_indices} serves as a lookup table.}
   \includegraphics[width=1\linewidth]{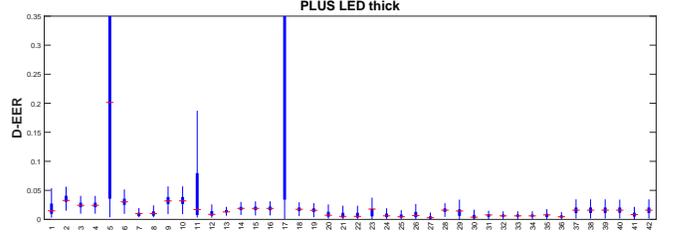}
\label{fig:boxplot_all}
\end{figure}

\begin{figure}
   \includegraphics[width=1\linewidth]{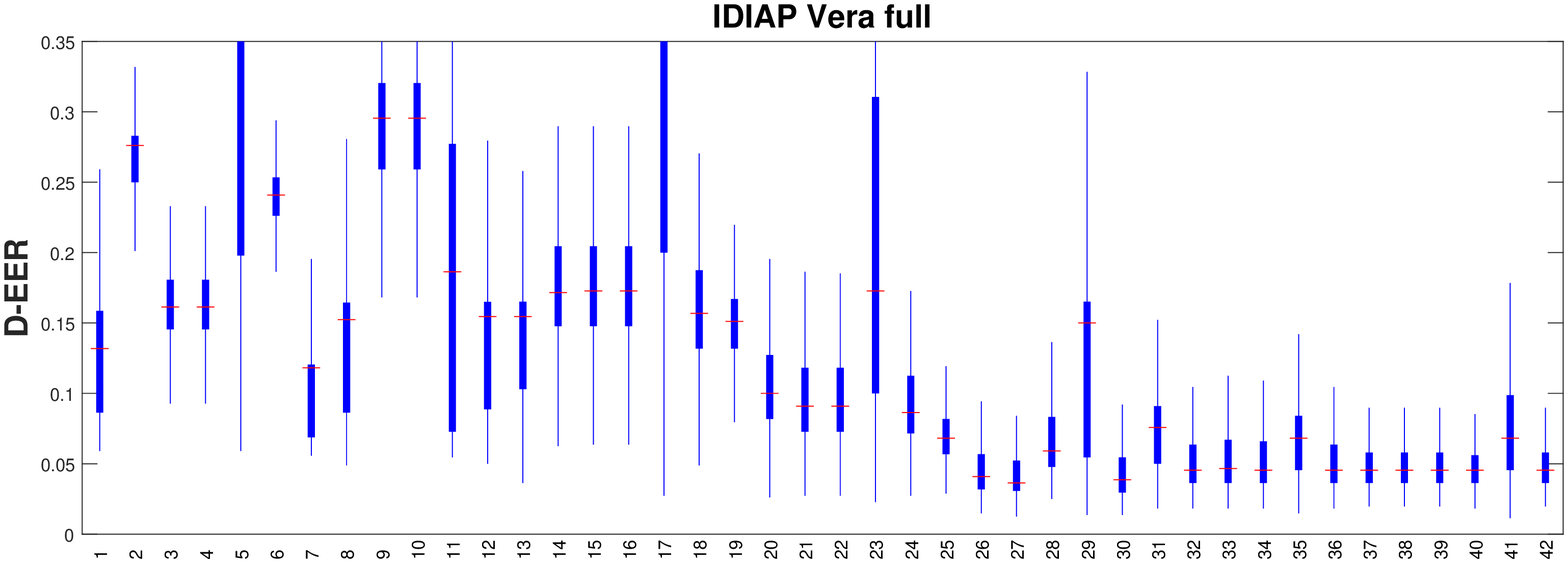}
\end{figure}
\begin{figure}
   \includegraphics[width=1\linewidth]{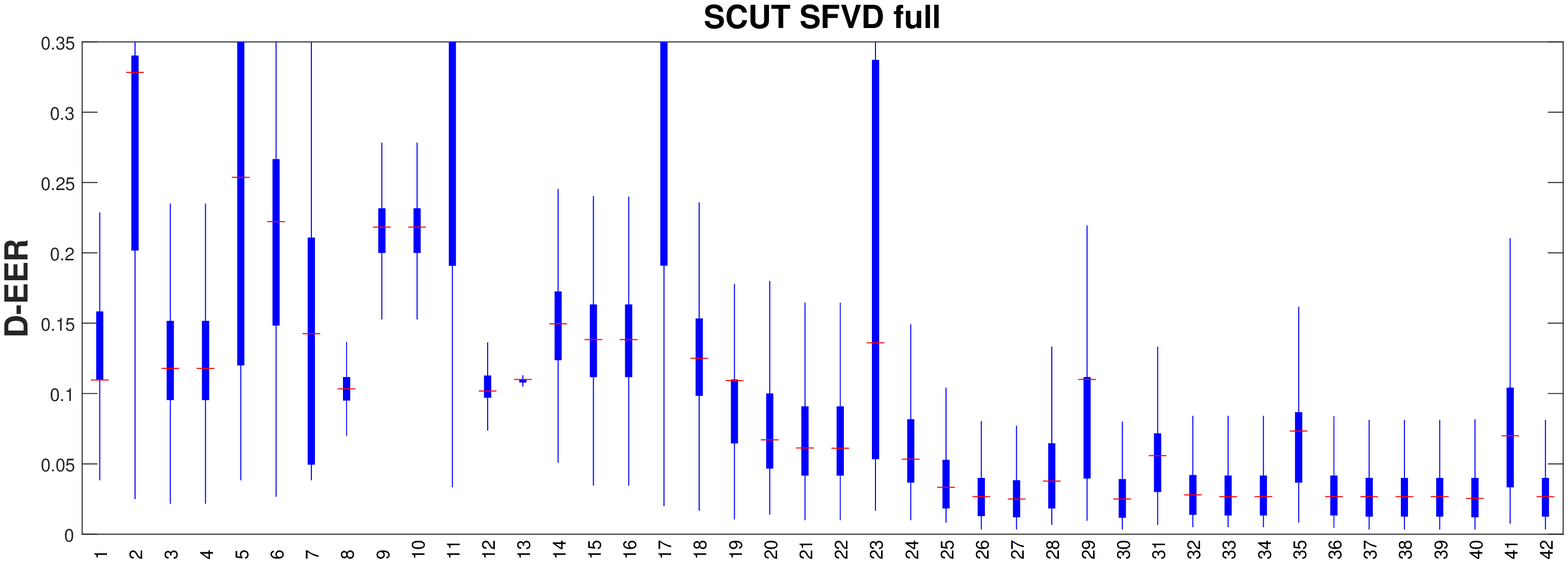}

\end{figure}

Since the SCUT database has a lot more samples than the other databases, the experiments in this section only considered comparisons of samples with IDs 1 and 2 per subject in order to keep the experiments computationally feasible in a reasonable amount of time. In order to split the comparison scores into train and test data, 2-fold cross validation is used. The best working constellation per database can be seen in in table \ref{tbl:PAD}. It is worth noting however that for the PLUS LED and PLUS Laser database often multiple method constellations would be eligible while still achieving 00.00\% D-EER. Hence, the reported constellations are chosen to require as few methods as possible and preferably use a computationally inexpensive norm-fusion combination. 
Each boxplot in figure \ref{fig:boxplot_all} represents 16\,383 method constellations for a particular normalization and fusion method, evaluated on a certain database. Every number on the x-axis corresponds to a certain norm-fusion-combination which can be deciphered using the lookup table \ref{tab:boxplot_indices}. One particular mode was chosen for every database since the overall trend is roughly the same for every sub-database. One can see that the score level fusion - presentation attack detection works well on the PLUS data. This observatuion coincides with the threat analysis, where the PLUS attacks had a hard time deceiving the keypoint, texture and minutiae recognition schemes, therefore all contributing valuable information for the PAD. The three peaks that can be seen at 5, 11 and 17 correspond to simple fusion rules together with unit-length norm. Interestingly though does the best working constellation for the IDIAP Vera full database include the unit-length norm together with the Fisher linear discriminant classifier. The overall trend that can be seen when looking at the IDIAP and SCUT boxplots indicates that the simpler fusion rules (more on the left side) are not so often a good choice, while the more complex fusion schemes (more on the right) often yield reasonable D-EERs.


\begin{table*}[t]
\caption{ Lookup table for the boxplot diagrams. The number on the x axis therefore indicates the norm-fusion combination that corresponding boxplot represents.  }
\centering
\begin{tabular}{| c | c | c | c | c | c |c |c |}
 \hline 
 & max-rule & min-rule & sum-rule & weighted-sum & svm-lin & svm-rbf  & lda-fusion \\
 \hline 
no-norm             & 1 & 7 & 13  & 19 & 25 & 31 & 37 \\
 \hline 
min-max-norm        & 2 & 8 & 14 & 20 & 26 & 32 & 38 \\
 \hline 
z-norm              & 3 & 9 & 15 & 21 & 27 & 33 & 39 \\
 \hline 
tanh-norm           & 4 & 10 & 16 & 22 & 28 & 34 & 40 \\
 \hline 
unit-length-norm    & 5 & 11 & 17 & 23 & 29 & 35 & 41 \\
 \hline 
rhe-norm            & 6 & 12 & 18 & 24 & 30 & 36 & 42 \\
 \hline
\end{tabular}
\label{tab:boxplot_indices}
\end{table*}

\section{Summary}
\label{sec:summary}

This study conducted an extensive threat analysis including three publicly available finger vein databases and one private dorsal hand vein database. The threat analysis was carried out using 14 vein recognition schemes that can be categorized into four types of algorithms based on what type of feature they extract. Experimental results show that all three finger vein databases pose a threat to most of the binarized vessel network based methods, while algorithms from the other categories behave more in-homogeneous. The general purpose keypoint scheme SIFT tends be very resistant against the PLUS attacks, while the other two finger vein databases seem to pose a threat, having IAPMRs ranging from 14\%\,-\,44\%. The other general purpose keypoint recognition scheme, SURF, seems to be very unimpressed by the presented attacks overall. Similar to the SIFT recognition scheme, the texture based methods LBP and CNN tend to be only susceptible to the IDIAP and SCUT attack samples. Minutiae based methods as well as the DTFPM keypoint scheme show minor (~5\%\,-\,16\% IAPMR) receptiveness for the PLUS attacks and higher (21\% and above) for the IDIAP and SCUT attacks. The second part of this research tries to use the in-homogeneity from the recognition schemes in order to perform presentation attack detection. To do so, comparison scores from different recognition schemes are combined using a range of different score level fusion techniques. After exhaustive cross-combination of all recognition schemes together with normalization and fusion schemes, the best working method constellations show indeed that presentation attack detection can be achieved.

\bibliographystyle{IEEEtran}
\bibliography{lit}

\end{document}